\documentclass{article}

\usepackage{PRIMEarxiv}

\usepackage[utf8]{inputenc} 
\usepackage[T1]{fontenc}    
\usepackage{hyperref}       
\usepackage{url}            
\usepackage{booktabs}       
\usepackage{amsfonts}       
\usepackage{nicefrac}       
\usepackage{microtype}      
\usepackage{lipsum}
\usepackage{fancyhdr}       
\usepackage{graphicx}       
\usepackage{subfig}
\usepackage[numbers]{natbib}
\graphicspath{{media/}}     

\title{The NPC AI of \textit{The Last of Us}: A case study
}

\author{
  Harsh Panwar \\
  School of ELectronic Engineering and Computer Science \\
  Queen Mary University of London \\
  London\\
  \texttt{h.panwar@se21.qmul.ac.uk} \\
}

\begin{document}
\maketitle

\begin{abstract}
\noindent \rule{\linewidth}{.5pt}
\textbf{Abstract:} \textit{The Last of Us} is game focused on stealth, companionship and strategy. The game is based in a lonely world after the pandemic and thus it needs AI companions to gain the interest of players. There are three main NPCs the game has - Infected, Human enemy and Buddy AIs. This case study talks about the challenges in front of the developers to create AI for these NPCs and the AI techniques they used to solve them. It also compares the challenges and approach with similar industry-leading games. 
\noindent \rule{\linewidth}{.5pt}
\end{abstract}


\section{Introduction}
\textit{The last of us} is a third-person shooter (TPS) action-adventure made by \textit{Naughty Dog} and distributed by \textit{Sony Computer Entertainment} developed majorly for \textit{PlayStation 3} and later on remastered for \textit{PlayStation 4} in 2014 \cite{dog2013last}. Since it's release the game has received amazing reviews by game developer critics \cite{hrej.cz} \cite{ry_2013} \cite{idnes.cz_2013} as well as by the gaming community and is considered as the best game of the decade as per Metacritic \cite{harradence_2020}. 

The game is set in post-apocalyptic America after the parasitic \textit{Cordyceps} fungus \cite{stark_2022} has wiped out majority of the humanity as we know it and divided the entire world into the infected and the survivors. In the nature this species of fungus \cite{evans1982cordyceps} can be seen attacking on the ants and taking control of their brains \cite{hywel1996cordyceps} forcing the ants to lose control and become a useless creature with only one job left - become host for the fungus, generating a massive sprout which eventually shoots out of their head and infect others eventually. Inspired by this natural phenomena, the creators of \textit{The last of Us} thought of a scenario where a similar fungus affected the human body. 

This paper is further organized into various sections where Section 2:\nameref{sec2} summarizes the various challenges faced by developers while creating the NPC AI. Section 3:\nameref{sec3} discusses the Skills and Behaviours AI system used in the game, Section 4:\nameref{sec4} presents the AI perception used and compares it with other games. Section 5:\nameref{sec5} presents the AI technique for follow positions. Section 6:\nameref{sec6} presents the AI technique used for cover and parts. Finally, Section 7:\nameref{sec7} concludes this paper.


\section{Challenges} \label{sec2}
There are mainly two types of people left after the pandemic has affected the entire world in \textit{The Last of Us} - The Infected and the Survivors. The pandemic in \textit{The Last of Us} has left most of the population disfigured and aggressive and those who have been overcome by the fungus are declared as Infected. The infected lacks any humanly characteristics such as compassion or self-preservation and are mainly driven by the instincts of the fungus. Then there are the survivors who survived the pandemic and now lives in quarantine zones and are at constant threats of the Infected and other predatory survivors known as Hunters. And there are also half a dozen of survivors which are companions of Joel, the main character, and help him travel in the lonely world. The developers had to majorly work on these three groups - The Infected, The Human Enemy and the Buddy companions. The developers chose to develop them using AI as such a system creates an enviornment closest to the real world. The players want the non-player characters (NPCs) in the game to be intelligent and just a simple glitch like an NPC running into a wall or trying to hide in plain sight can result in player's disinterest in the game. In a study done by Tremblay et. al. \cite{tremblay2013adaptive} they found that the adaptive AI companion has more influence over the player's experience and it is a step forward in developing meaningful and engaging games. But each of these groups have there own challenges when it comes to developing the AI for them and they are discussed below in brief:

\begin{figure}%
    \centering
    \subfloat[\centering Ellie, a 14-year old teenage girl who accompanies Joel in most part of the game and is dependent on him.]{{\includegraphics[width=7cm]{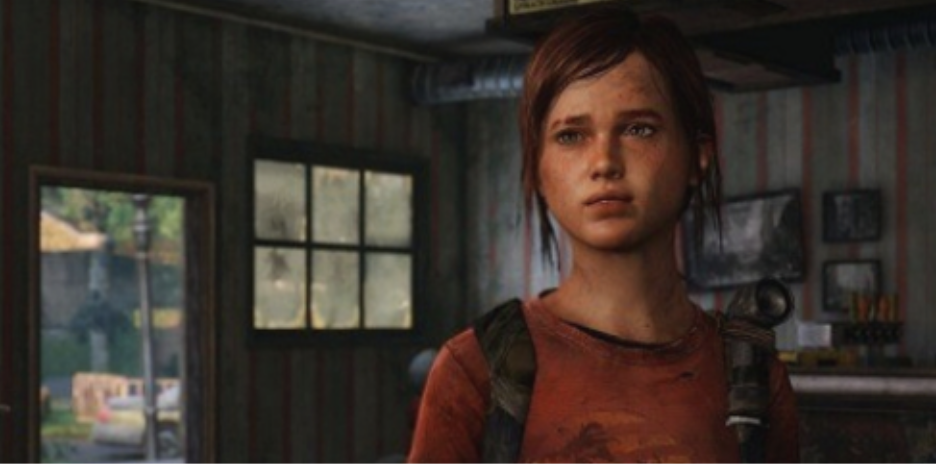} }}%
    \qquad
    \subfloat[\centering Elizabeth, who reveals special abilities and has a complex background story. ]{{\includegraphics[width=7cm]{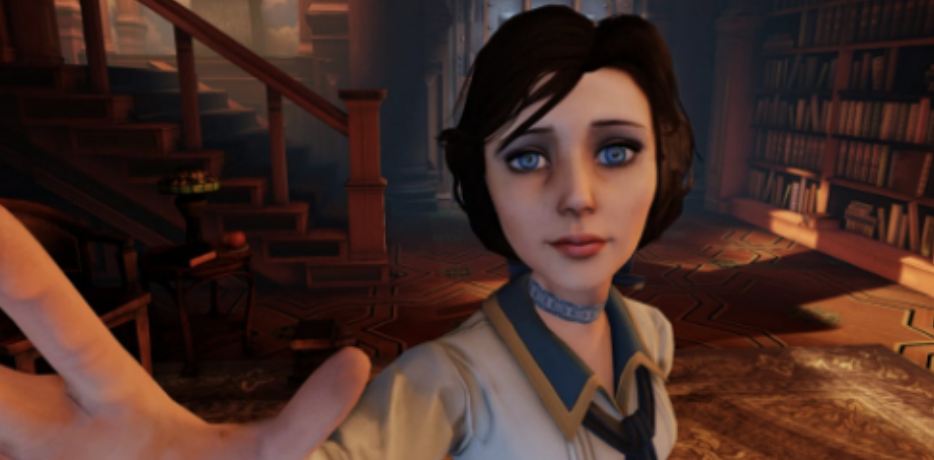} }}%
    \caption{Screenshot from the games showing the companion AIs in (a) \textit{The Last of Us} and (b) \textit{Bioshock: Infinite}}%
    \label{fig:example}%
\end{figure}

\subsection{The Infected}
One basic understanding of AI in Games is to make the NPCs intelligent and smarter. This simply can't be done for the Infected AI as they have been made dumb by the fungus. This was one of the biggest challenges for the AI Developers in \textit{The Lat of Us} as they had to make them seem chaotic and alien unlike Hunters which work in groups and coordinate with each other. Our understanding of intelligence is more closely related to humanly behaviours such as togetherness, coordination, expressions etc. which were not allowed to be use for the Infected. \\

\begin{figure}
    \centering
    \includegraphics[width=14cm]{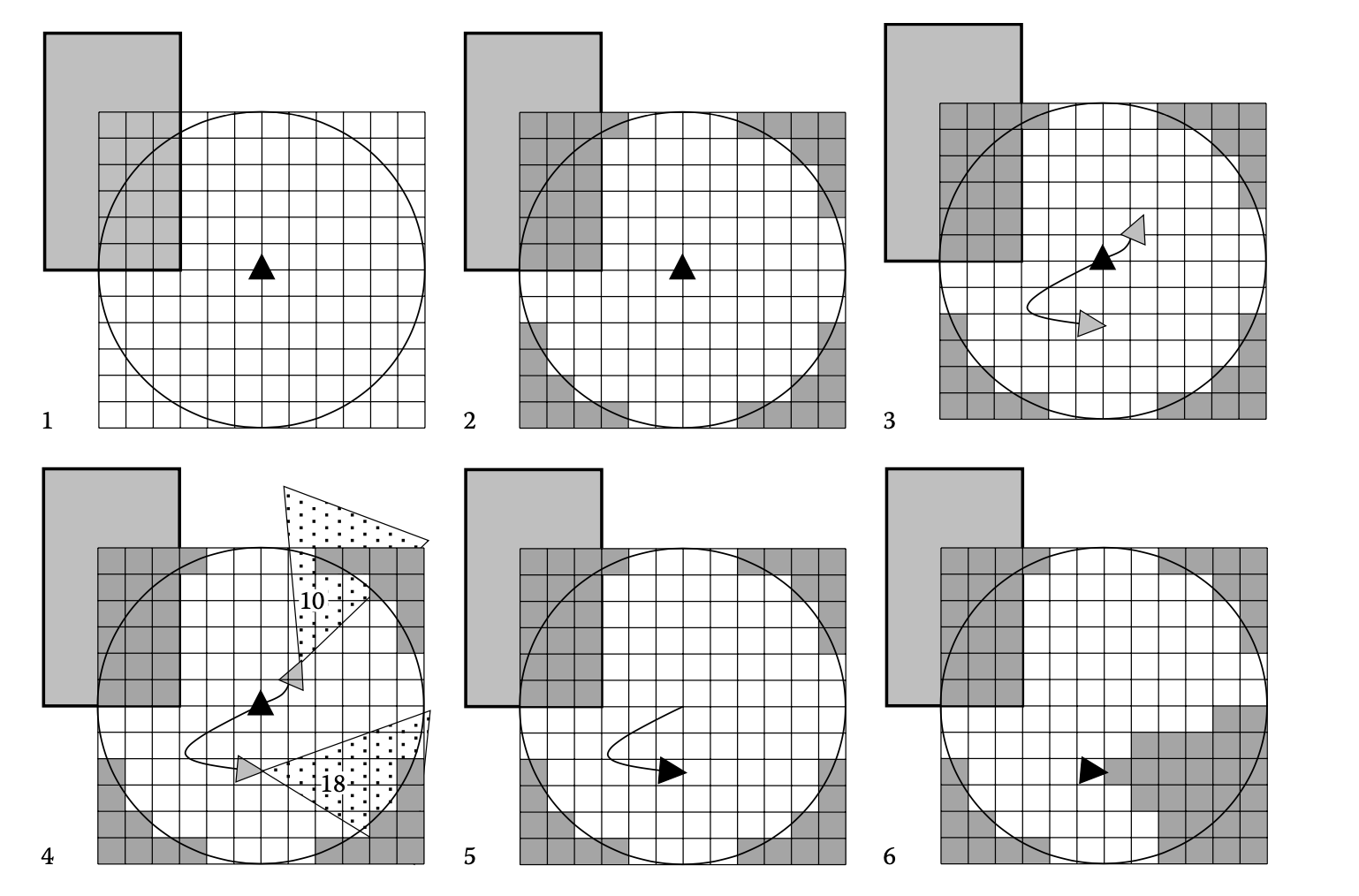}
    \caption{\textit{infected-canvass} behaviour}
    \label{fig:Search_SKill}
\end{figure}

The Infected also because of the fungus were left to be sightless and had to rely mostly on hearing. This made it even more difficult to make them seem more real as otherwise it'll become very easy to tackle them and the player could lose interest in the game. Just like in the actual world if you become blind or lose one of the 5 senses then the other senses for example hearing becomes stronger \cite{Karns9626}, the developers had to make sure that their AI is really smart when it comes to actions and reactions using the ability of hearing.

\subsection{Human enemy}
The primary challenge while creating the Human AI was how to make an NPC so impactful that killing that NPC will have a effect on the player which will make the game interesting. Unlike in \textit{Alien:Isolation} where the Xenomorph, primary antagonist, has two main AIs - main control itself using behavior trees which is very common in AAA games and an AI-director just like \textit{Left 4 Dead} \cite{kitson2009left} which mainly manages the pacing of the game and the main focus was on Sense, Menace and Speed \cite{tommy}. The Human Enemy AI in \textit{The Last of Us} had to be focused mainly on Cover, Stealth and Flanking to make them look more like real humans and work in groups to give a feel of fighing against actual humans. 

\subsection{Buddy companions}
The companion AI is one of the key factor in making a game like \textit{The Last of Us} interesting since in a post-apocalyptic world which is full of Infected alienated creatures it's essential that the only friend we have is intelligent and also compassionate. In an online survey done by Emmerich et. al. \cite{10.1145/3242671.3242709} on players they found out that most players prefer to have a companion and the most important charateristics they look in the companion are - Personality $(M = 3.82)$, Context Sensitivity $(M = 3.53)$ and Story Significance $(M = 3.70)$ on a Mean scale: $0-4$. \textit{Elizabeth} is a companion AI in \textit{Bioshock: Infinte} \cite{bio2k} but is very different from \textit{Ellie} in a way that \textit{Elizabeth} has special abilities to help the player while \textit{Ellie} is more dependent on the player. Since in \textit{The Last of Us} the characters lacks any superpower or mysterious abilities, \textit{Ellie} who's just a 14-year old teenager is very difficult to be developed as someone who the player will find useful and not a liability who has to be escorted. Making a character that we as a player cared for was the key to the success of the game.

The pandemic in \textit{The Last of Us} has left most of the population disfigured and aggressive and those who have been overcome by the fungus are declared as Infected. Although the AI for the Infected and the Hunters are similar, the developers have made the architecture modular which helps to make changes to the decision-making logic which makes the Infected feel fundamentally different than the Hunters \cite{botta2019infected}.

\section{Skills and Behaviours} \label{sec3}
The AI System used in \textit{The Last of Us} mainly comprises of high-level decision logic known as the skills and low-level capabilities known as behaviours. The skills are used to decide what the character should do while the behaviours are used to implement those decisions. This sort of model helps in re-using the same low-level behaviours for multiple high-level characteristics. The skills are stacked in a priority queue which tells us which skills can interrupt other skills. 
The examples for skills for the infected are - \textit{chase, search, follow, sleep, wander, etc.} and for human enemy the skills are - \textit{panic, advance, melee, gun combat, hide, investigate, scripted and flank}. Since the skills are high-level they don't change the animations or pathfinding systems. These changes are instead made by behaviour object using a behaviour stack. Examples of behaviour are MoveToLocation, StandAndShoot and TakeCover.

Most of the skills are common between the different characters for ease of adding new characters at any stage of development. There are 4 types of Infected - \textit{Runner, Stalker, Clicker and Bloater} and only two skills are unique to single characters - \textit{ambush} skill by the \textit{stalker} and \textit{throw} skill by the \textit{bloater}. Now we will explain some of the major skills in detail:
\subsection{Search Skill}
The hunters are more intelligent beings and can communicate with each other to gather knowledge about the player's position. The infected on the other hand lacks this and also can hardly see. So a requirement was there for a special search skill for the infected. This skill is in low priority compared to the \textit{chase} skill and thus becomes valid when the infected loses track of the player during \textit{chase} phase. The search by the infected should look unplanned and random while also covering a large amount of area. The search points \cite{straatman2006dynamic} can be anywhere on the navigation mesh and hence the search area scan be of arbitary size and shape. To solve this \textit{infected-canvass} behaviour was developed as shown in fig \ref{fig:Search_SKill} and the steps are explained below:
\begin{enumerate}
    \item With the Infected in the center we place a logical grid on the covered area by the \textit{canvass radius}.
    \item Any obstacle that comes in the way and the entire area outside is fixed as \textit{seen} and the rest of the cells needs to be checked.
    \item Multiple possible animations are given to the behaviour by the invoking skills. Location and orientation of the Infected after these animations is looked at.
    \item The number of unseen cells is computed for every animation. 
    \item We play the most desirable animation generally by counting the number of unseen cells and ignore the animations that were played recently.
    \item After the completion of the wedges they are marked as seen and the process is repeated from step 3. 
\end{enumerate}

\section{AI Perception} \label{sec4}
\begin{figure}
    \centering
    \includegraphics[width=14cm]{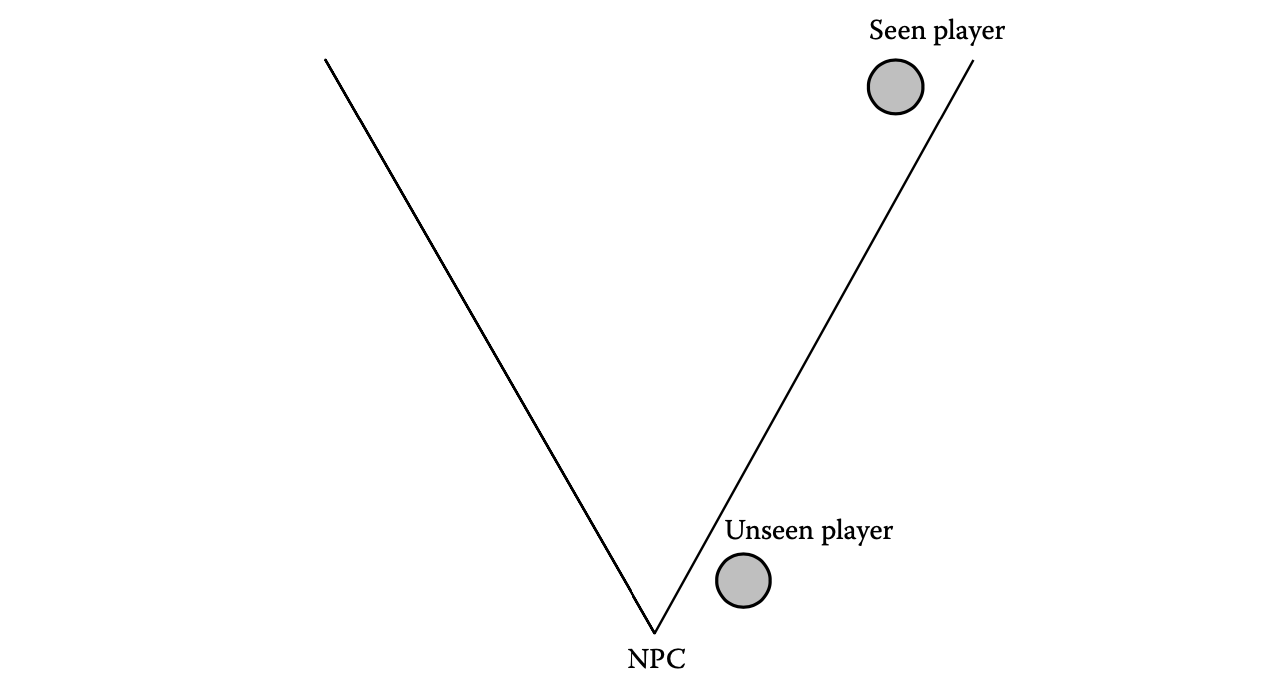}
    \caption{The initial vision cone developed for \textit{The Last of Us} which was unable to look at players standing right next to the NPC.}
    \label{fig:Vision_1}
\end{figure}

\begin{figure}%
    \centering
    \subfloat[\centering Four Vision cone used for the Xenomorph in the \textit{Alien: Isolation} - \textit{normal} (Blue), \textit{focused} (Red), \textit{peripheral} (Green) and \textit{close} (Pink). ]{{\includegraphics[width=7cm]{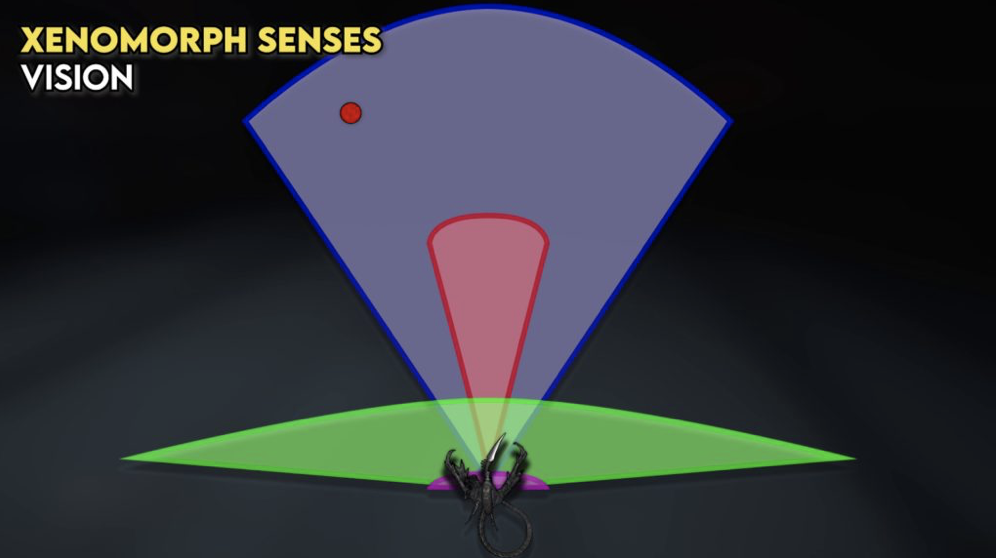} }}%
    \qquad
    \subfloat[\centering The final complex vision cone used in \textit{The last of Us} which is easily able to see the player in close proximity. ]{{\includegraphics[width=7cm]{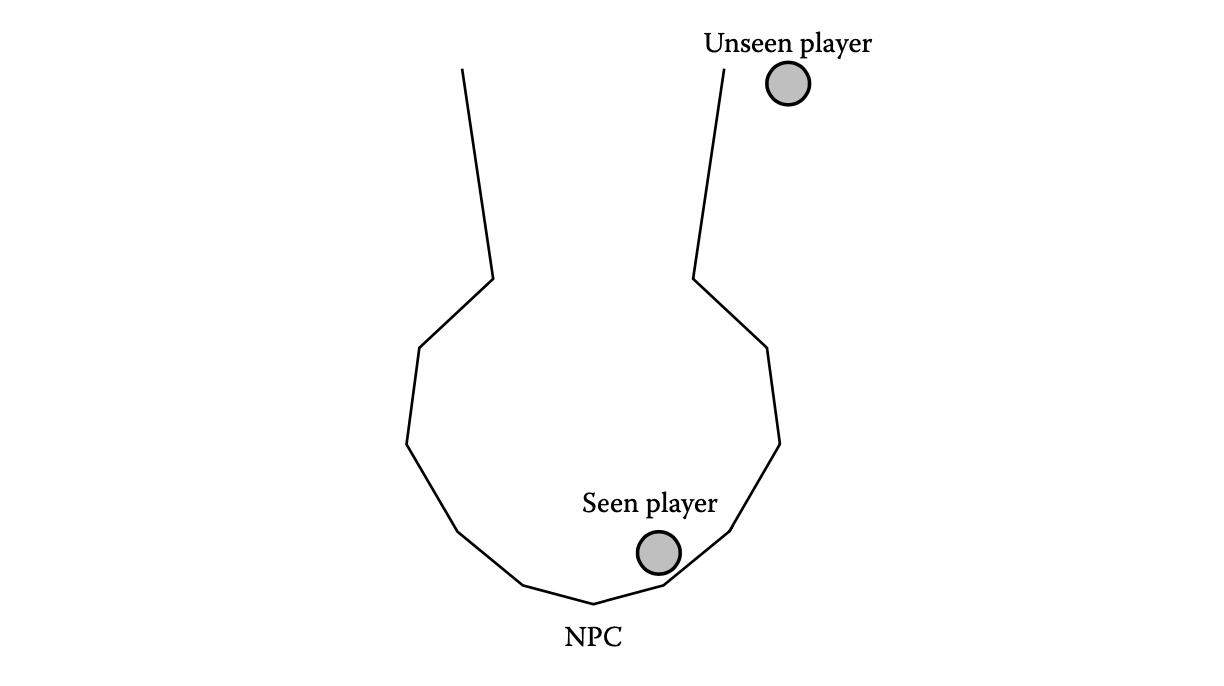} }}%
    \caption{Screenshot from the games showing the companion AIs in (a) \textit{The Last of Us} and (b) \textit{Bioshock: Infinite}}%
    \label{fig:Vision_23}%
\end{figure}

AI perception is one of the key factor in determining the success of any Game AI. Especially in a game like \textit{The Last of Us} which is heavily based on stealth it's important to know the position of other characters. The developers initially decided to use Vision Cone which is a very common way of visualising how the enemy sees the world and has been used in games like \textit{The Uncharted} \cite{dog2015art} and \textit{Alien: Isolation}. As seen in Fig. \ref{fig:Vision_1} using a Vision Cone is very effective in finding player which are at a distance but fails to register player which is at close proximity to the NPC or standing right next to the NPC. This lead to the NPC acting very dumb and unaware of the surrounding which made it easy to full them and lead to the players losing interest in the game. One way to solve this is to have multiple Vision cone similar to the ones used in the \textit{Alien: Isolation} Xenomorph AI\cite{tommy}. It uses 4 different view cones namely - \textit{normal, focused, peripheral and close} as seen in Fig. \ref{fig:Vision_23}(a). The developers of \textit{The Last of Us} came up with an even better and complex form of Vision cone with a simple rule that the angle of view is inversely proportional to the distance between the NPC and the player as seen in the Fig. \ref{fig:Vision_23}(b).

\section{Follow Positions} \label{sec5}
\begin{figure}
    \centering
    \includegraphics[width=14cm]{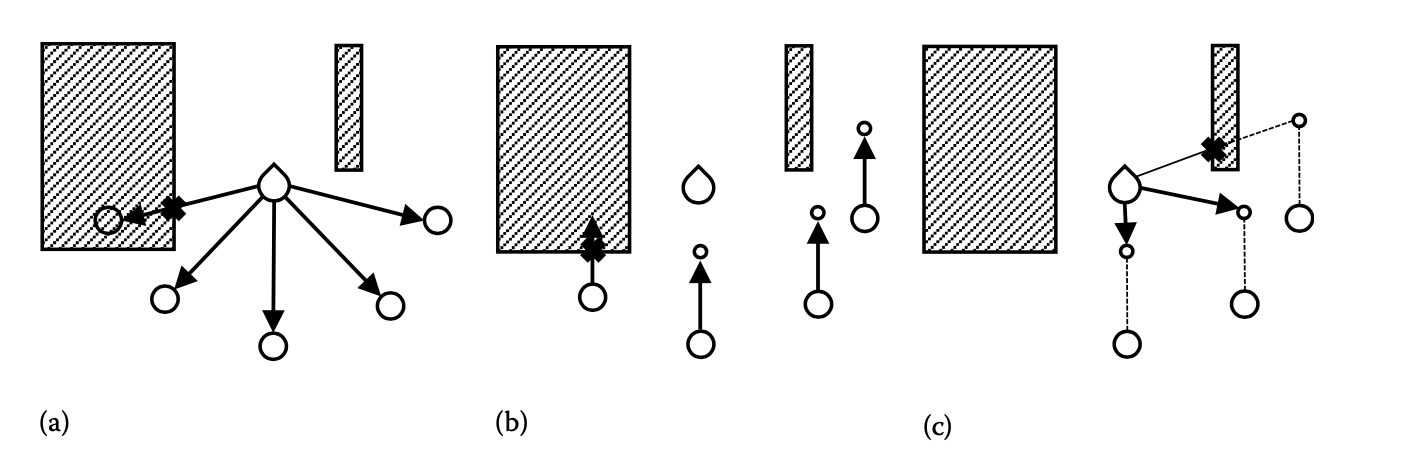}
    \caption{Pathfinding raycasts for follow positions (a) for generating candidate positions (b) for checking forward location (c) for checking future position}
    \label{fig:follow_SKill}
\end{figure}

It's essential for the Buddy AI, \textit{Ellie}, to follow the player and stay close to him to not be responsible for raising any alerts to the enemy or even if something bad happened it can be related to the players on mistake since they both were close to each other and the player shouldn't have been that exposed in the first place. To bring this in the game a follow system was developed which generates a number of candidate follow positions which are evaluated for quality. This generation was done by casting raycasts as seen in Fig. \ref{fig:follow_SKill}. 

\begin{itemize}
    \item First the raycasts are sent from the player to the follow region to make sure of a clear path between the player and the Buddy AI. Every raycasts that sucessfully reaches the follow region is considered as a candidate. See fig. \ref{fig:follow_SKill}(a)
    \item Secondly new set of rays are shooted forward from every candidate to avoid any walls. See fig. \ref{fig:follow_SKill}(b)
    \item And finally from the players position rays are cast to each forward position to ensure obstacle free-movement. See fig. \ref{fig:follow_SKill}(c) 
\end{itemize}

\begin{figure}
    \centering
    \includegraphics[width=14cm]{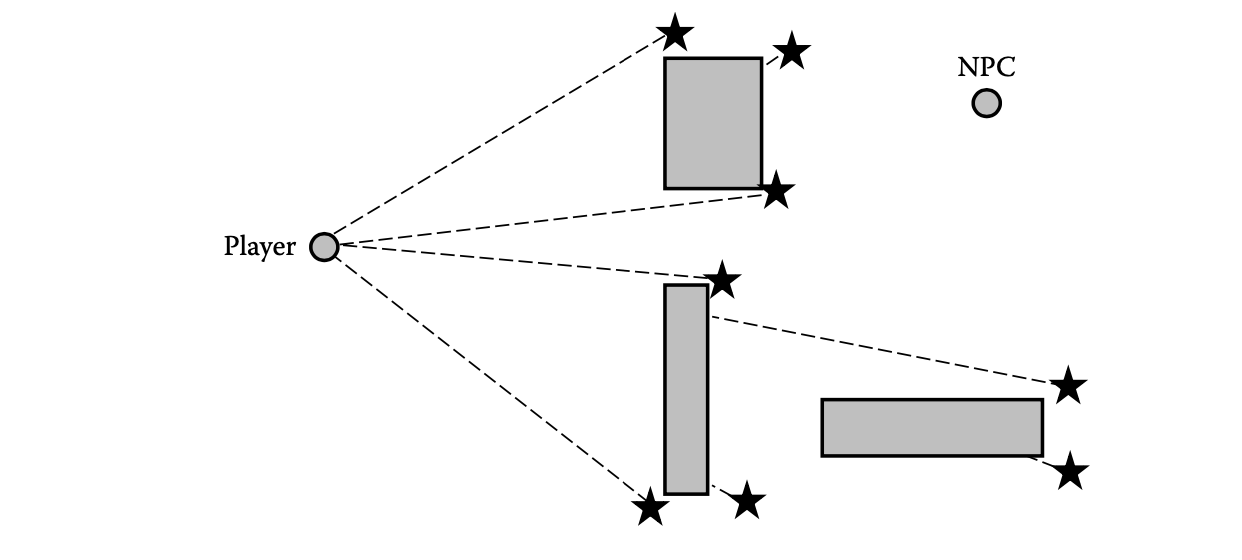}
    \caption{Raycasts being generated for coverposts.}
    \label{fig:cover_1}
\end{figure}

\begin{figure}
    \centering
    \includegraphics[width=14cm]{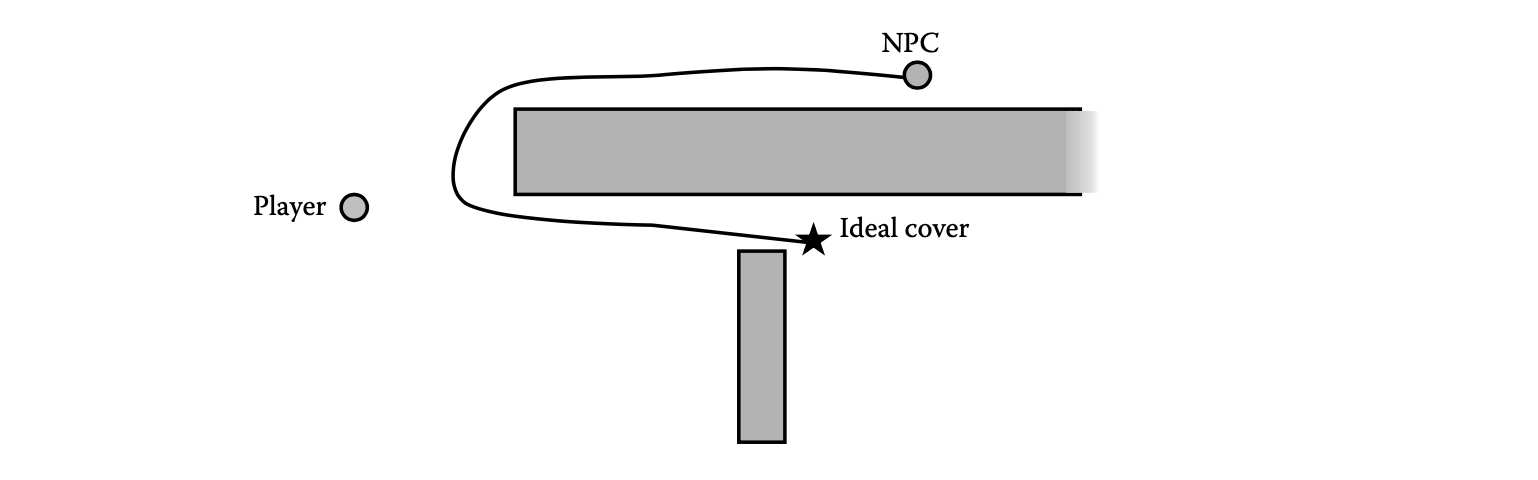}
    \caption{Some of the best cover could have a path which leads to the player before reaching the cover.}
    \label{fig:cover_2}
\end{figure}

\begin{figure}
    \centering
    \includegraphics[width=14cm]{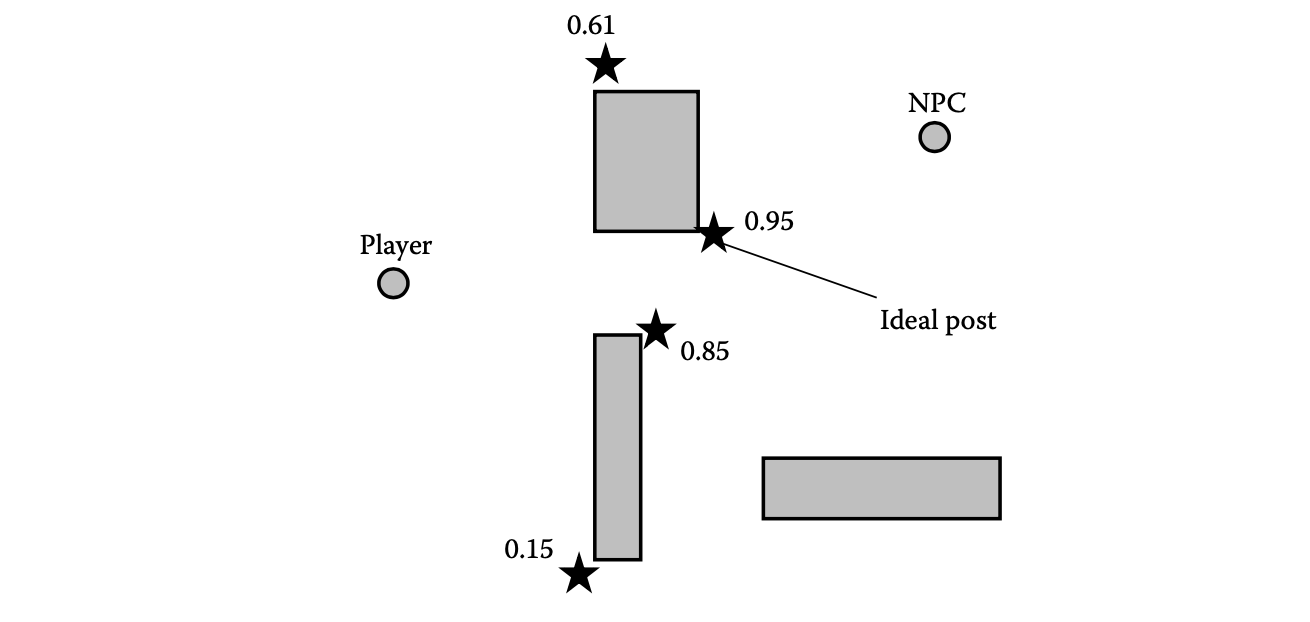}
    \caption{Ranking for each posts calculated based on AI criterion written in LISP.}
    \label{fig:cover_3}
\end{figure}

\section{Cover and Posts} \label{sec6}
\textit{The Last of Us} is a cover-based TPS \cite{dyckhoff2019ellie} and the main focus is on stealth based attacking and strategising each move. For the AI to look and feel human-like it's a very important decision of where to stand or take cover and is a fairly complex problem. For establishing proper covers we first need to recognise a set of potential locations which we'll call \textit{posts}. 
There are two different kind of \textit{posts} - cover and open. The cover \textit{posts} were formed around the NPC's location while the open posts were around the player. Each NPC would be assigned 20 cover \textit{posts} and in each frame 160 raycasts were thrown to them. The \textit{posts} where every ray was rejected out of the 4 were rejected as shown in Fig. \ref{fig:cover_1} 
The posts an NPC should use would be determined by different criterion written in LISP \cite{steele1990common} using \textit{post selectors}. After the development of the game there were 17 different \textit{post selectors} with the most important one being \textit{ai-criterion-static-pathfind-not-near-player} \cite{mcintosh2015human}.
This was used when the path to the post required the human enemy to cross the player as shown in Fig. \ref{fig:cover_2}. This wouldn't make sense and will lead to the player easily killing the human enemy and making the difficulty easier. Each of the criterion would give some value in float numbers and the product of the criteria for a given post selector and a given post in a value which would be used as the rating for that post. Then all the \textit{posts} would be ranked accordingly and the one with the highest score would be chosen for that NPC as shown in Fig. \ref{fig:cover_3}.

\section{Conclusion} \label{sec7}
The Last of Us was a complete success and received love from the players mostly because of how close to reality the game tried to go thanks to it's AI. The game was in a sense first of it's kind where the difficulty of the game didn't lie in making the characters hard to kill but by increasing the co-ordination between the NPCs and thus increasing the intelligence in the game characters. Through the AI used in Ellie the players were able to create an actual relationship with her and each and every gesture performed by her made a meaning to the player. This leads us to question what we consider as human-like and can we make a companion AI in video games that will feel like an actual human being. The level of determination that the developers of \textit{The Last of Us} went to can be seen when they talk about "her personal space" and "her choices" in one of the papers they have written on the character. The developers also avoided cheating in most ways with things like only using Teleportation when extremely necessary and in a way that the player don't realize makes us understand that why the game feels so engaging to the players.

\bibliographystyle{unsrtnat}  
\bibliography{references}

\end{document}